\documentclass{article} 
\usepackage{tencent_ailab_tech_report}
\usepackage[colorlinks = true,
            linkcolor = blue,
            urlcolor  = blue,
            citecolor = blue,
            anchorcolor = blue]{hyperref}

\usepackage{graphicx}
\usepackage{caption}
\usepackage{subcaption}
\usepackage{booktabs} 

\usepackage{fancyvrb}
\usepackage{listings}
\usepackage{xcolor}

\lstdefinelanguage{json}{
    basicstyle=\ttfamily\small,
    numbers=left,
    numberstyle=\tiny,
    stepnumber=1,
    numbersep=5pt,
    showstringspaces=false,
    breaklines=true,
    frame=lines,
    backgroundcolor=\color{gray!10},
    literate=
     *{0}{{{\color{blue}0}}}{1}
      {1}{{{\color{blue}1}}}{1}
      {2}{{{\color{blue}2}}}{1}
      {3}{{{\color{blue}3}}}{1}
      {4}{{{\color{blue}4}}}{1}
      {5}{{{\color{blue}5}}}{1}
      {6}{{{\color{blue}6}}}{1}
      {7}{{{\color{blue}7}}}{1}
      {8}{{{\color{blue}8}}}{1}
      {9}{{{\color{blue}9}}}{1}
      {:}{{{\color{red}:}}}{1}
      {,}{{{\color{red},}}}{1}
      {"}{{{\color{orange}"}}}{1}
      {[}{{{\color{green}[}}}{1}
      {]}{{{\color{green}]}}}{1}
      {\{}{{{\color{green}\{}}}{1}
      {\}}{{{\color{green}\}}}}{1}
}

\usepackage{CJKutf8}
\usepackage{natbib}
\usepackage{microtype}
\usepackage{hyperref}
\usepackage{url}
\usepackage{booktabs}
\usepackage{enumitem}
\usepackage{multicol}
\usepackage{multirow}
\usepackage{CJKutf8}
\usepackage{amsmath}
\usepackage{siunitx}
\usepackage{floatflt}
\usepackage{wrapfig}
\usepackage{graphicx}
\usepackage{booktabs}
\usepackage{wrapfig}
\usepackage{authblk}
\usepackage{lipsum}
\usepackage{dsfont}
\usepackage{bbm}
\usepackage{algorithm}
\usepackage{algorithmicx}
\usepackage{algpseudocode}
\usepackage{microtype}
\usepackage{graphicx}
\usepackage{subcaption}
\usepackage{multirow}
\usepackage{booktabs} 
\usepackage{pifont}  
\usepackage{graphicx}  
\usepackage{subcaption} 
\usepackage{hyperref}
\usepackage[table]{xcolor}
\usepackage[most]{tcolorbox}
\newtcolorbox{alprompt}[1]{
        boxrule = 1pt,
        fontupper = \small\tt,
        fonttitle = \bf\color{black},
        arc = 2pt,
        rounded corners,
        colframe = black,
        colbacktitle = white!97!yellow,
        colback = white!97!yellow,
        title = #1,
}
\definecolor{darkgreen}{rgb}{0.0, 0.5, 0.0}
\definecolor{darkgray}{gray}{0.4}
\definecolor{maroon}{rgb}{0.5, 0.0, 0.0}
\definecolor{navy}{rgb}{0.0, 0.0, 0.5}
\definecolor{teal}{rgb}{0.0, 0.5, 0.5}

\definecolor{deepblue}{RGB}{41, 128, 185}

\definecolor{mylightgreen}{RGB}{144,238,144}
\definecolor{mylightblue}{RGB}{173,216,230}

\definecolor{outerboxcolor}{gray}{0.90} 
\definecolor{innerboxcolor}{rgb}{1,1,1}

\definecolor{nred}{RGB}{196, 38, 11}
\definecolor{ngreen}{RGB}{18, 141, 21}
\definecolor{nblue}{RGB}{41, 52, 190}

\definecolor{SecPurple}{RGB}{238,223,246}   
\definecolor{SecOrange}{RGB}{255,235,218}   
\definecolor{secblue}{RGB}{220,235,255}   
\definecolor{green}{RGB}{220,245,225}   
\definecolor{secsecblue}{RGB}{245, 250, 255}

\usepackage{listings}
\usepackage{xcolor}
\usepackage{adjustbox}
\usepackage{makecell}
\usepackage[para,online,flushleft]{threeparttable}

\usepackage{wrapfig}
\usepackage{amssymb}
\usepackage{amsthm}

\theoremstyle{plain}

\theoremstyle{definition}

\theoremstyle{remark}

\usepackage{pifont}
\usepackage{tabularx}

\algnewcommand{\LeftComment}[1]{\Statex \(\triangleright\) #1}

\usepackage{array}
\usepackage{amsmath}
\usepackage{mathtools}
\usepackage{amsthm}
\usepackage{arydshln}
\usepackage[capitalize,noabbrev]{cleveref}
\usepackage{adjustbox} 
\usepackage{enumitem}
\usepackage{longtable}

\usepackage[textsize=tiny]{todonotes}

\newcommand{\truncated}{\textit{[...]} }

\makeatletter
\renewcommand{\@fnsymbol}[1]{\ensuremath{\ifcase#1\or \dag\or \ddag\or \S\or \P\or \|\or **\fi}}
\makeatother

\colmfinalcopy

\title{AzeroS: Extending LLM to Speech with Self-Generated Instruction-Free Tuning}

\hypersetup{
    pdftitle={AzeroS: Extending LLM to Speech with Self-Generated Instruction-Free Tuning},
    pdfauthor={Yiwen Shao, Wei Liu, Jiahong Li, Tianzi Wang, Kun Wei, Meng Yu, Dong Yu}
}

\begin{document}

\pagestyle{fancy}
\fancyhead{} 
\renewcommand{\headrulewidth}{1.5pt}

\ifcolmfinal
    \lhead{\raisebox{0.1cm}[0pt][0pt]{AzeroS}}
    \rhead{\includegraphics[height=0.45cm]{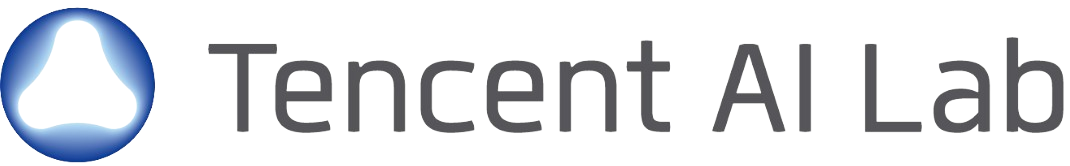}}
    \fancypagestyle{firstpage}{%
        \fancyhf{}%
        \fancyhead[R]{\includegraphics[height=0.45cm]{tencent_ai_lab_logo.png}}%
        \fancyfoot[C]{\thepage}%
        \renewcommand{\headrulewidth}{1.5pt}%
    }
    \thispagestyle{firstpage}
\else
    \lhead{Under review as a conference paper at COLM 2024}
    \rhead{}
\fi

\begin{tcolorbox}[
    colback=secsecblue, 
    colframe=secsecblue, 
    arc=10pt, 
    boxrule=0pt, 
    width=\textwidth, 
    left=15pt, right=15pt, top=15pt, bottom=15pt 
]

    {\Large \bfseries AzeroS: Extending LLM to Speech with Self-Generated Instruction-Free Tuning \par}
    
    \vspace{0.8em}
    
    \ifcolmfinal
        {\normalsize \bfseries 
        Yiwen Shao\textsuperscript{*\dag}, Wei Liu\textsuperscript{*}, Jiahong Li\textsuperscript{*}, Tianzi Wang, Kun Wei, Meng Yu, Dong Yu
        \par}
        
        \vspace{0.4em}
        
        {\small Tencent AI Lab \par} 
        {\footnotesize \color{darkgray} \ttfamily yiwenyshao@global.tencent.com \par}
        \vspace{0.2em}
        {\scriptsize \color{gray} \textit{\textsuperscript{*}Equal Contribution, \textsuperscript{\dag}Project Lead}}
    \else
        {\normalsize \bfseries Anonymous authors \par}
        \vspace{0.2em}
        {\normalsize Paper under double-blind review \par}
    \fi
    
    \vspace{1em}
    
    \noindent 
    {\small 
    \textbf{Abstract:} Extending large language models (LLMs) to the speech domain has recently gained significant attention. A typical approach connects a pretrained LLM with an audio encoder through a projection module and trains the resulting model on large-scale, task-specific instruction-tuning datasets. 
However, curating such instruction-tuning data for specific requirements is time-consuming, and models trained in this manner often generalize poorly to unseen tasks.
In this work, we first formulate that the strongest generalization of a speech--LLM is achieved when it is trained with \textbf{Self-Generated Instruction-Free Tuning (SIFT)}, in which supervision signals are generated by a frozen LLM using textual representations of speech as input. Our proposed SIFT paradigm eliminates the need for collecting task-specific question--answer pairs and yields the theoretically best generalization to unseen tasks.  
Building upon this paradigm, we introduce \textbf{AzeroS} (\textbf{A}uden \textbf{Zero}-instruction-tuned \textbf{S}peech-LLM), which is trained on speech--text pairs derived from publicly available corpora, including approximately 25k hours of speech with ASR transcripts and 3k hours of speech with paralinguistic labels. Built upon Qwen2.5-7B-Instruct, the model updates only two lightweight projection modules (23.8M parameters each), while keeping both the LLM and audio encoders frozen. Despite the minimal training cost and modest data scale, AzeroS achieves state-of-the-art performance on both semantic and paralinguistic benchmarks, including VoiceBench, AIR-Bench Foundation (Speech), and AIR-Bench Chat (Speech).
    }
    
    \vspace{1.5em}
    
    \noindent
    \ifcolmfinal
        \small 
        \textbf{Code:} \url{https://github.com/AudenAI/Auden/tree/main/examples/azeros} \\
        \textbf{Model:} \url{https://huggingface.co/AudenAI/azeros/tree/main} \quad
    \else
        \small \textit{Links omitted for blind review.}
    \fi

\end{tcolorbox}

\vspace{2em}

\begin{figure*}[h]
  \centering
  \includegraphics[width=0.9\linewidth]{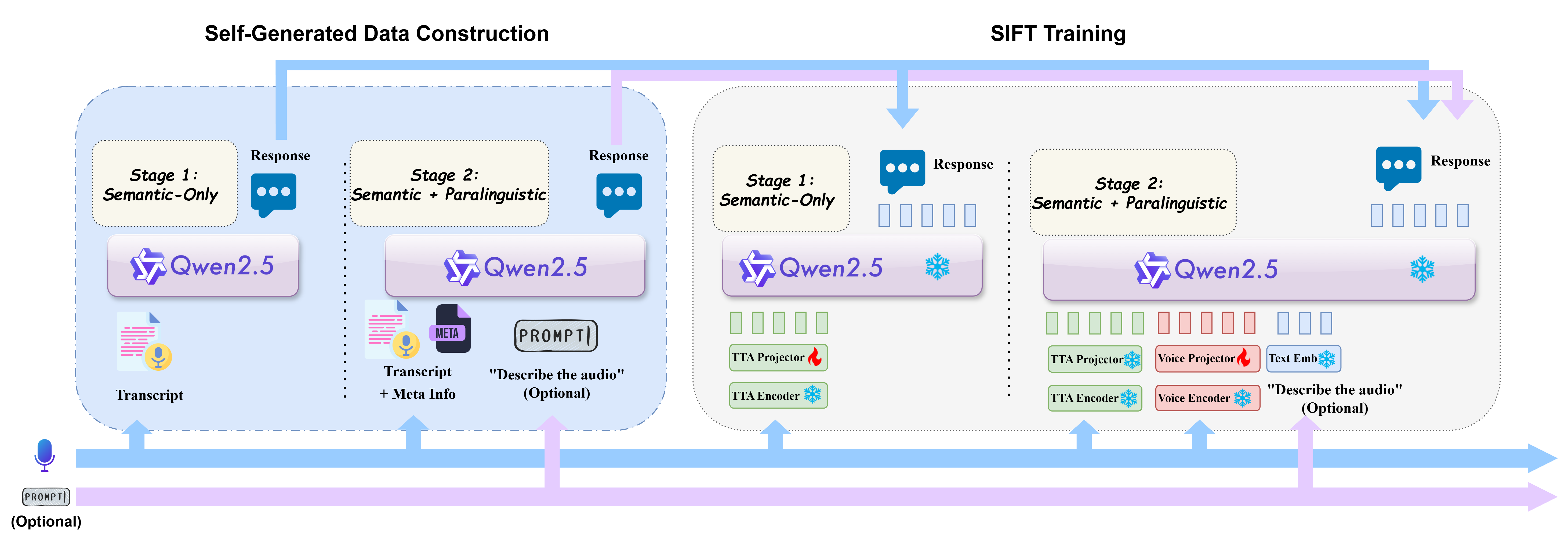}
  \caption{\small
Overview of the proposed AzeroS framework built upon a frozen Qwen2.5-7B-Instruct backbone.
The system couples a \textbf{Self-Generated Data Construction} pipeline (left) with a two-stage \textbf{Self-Generated Instruction-Free Tuning (SIFT)} training procedure (right).
In \textbf{Stage~1 (Semantic)}, the LLM generates targets from speech transcripts to train a projector on the semantic \textit{TTA}~\cite{auden_tta} encoder.
In \textbf{Stage~2 (Semantic + Paralinguistic)}, we augment inputs with metadata (e.g., emotion, gender, age); the resulting rich responses train a second projector on a paralinguistic \textit{Auden-Voice}~\cite{auden_voice} encoder.
Together, this progressive SIFT paradigm extends the model's capability from purely semantic to joint semantic–paralinguistic understanding without relying on task-specific instructions.
}
  \label{fig:overview}
\end{figure*}
\section{Introduction}
Large language models (LLMs)~\cite{mann2020language, du2021glm, hoffmann2022training, chowdhery2023palm, chiang2023vicuna, chung2024scaling,achiam2023gpt,touvron2023llama,bai2023qwen,yang2024qwen2,zhang2022opt,henighan2020scaling,kojima2022large,rae2021scaling,jiang2023mistral7b} have revolutionized the field of Natural Language Processing (NLP). Trained on internet-scale text data through next-token prediction, LLMs acquire vast world knowledge and exhibit powerful language understanding and reasoning capabilities. In recent years, extending LLMs to incorporate additional modalities has become a prominent research trend.  By inheriting LLMs' language generalization abilities, it is expected to develop unified multimodal LLMs (MLLM)~\cite{radford2021learning, liu2023visual,comanici2025gemini,yin2024survey} that can handle multiple tasks across various modalities using a single model.

Among all modalities, speech plays a fundamental role in human communication. A straightforward way to integrate it with LLMs is through a cascaded system.
The cascaded speech--LLM pipeline, which first applies automatic speech recognition (ASR) before text-based processing, suffers from error propagation and the loss of non-semantic acoustic information such as prosody and emotion. These limitations motivate the development of end-to-end architectures that process speech directly.

End-to-end speech--LLMs seamlessly connect a speech encoder with an LLM without intermediate text. Forwarding with speech representation mitigates error propagation and could potentially preserve those fine-grained, non-semantic cues. Existing approaches can be broadly categorized into three types~\cite{arora2025landscape}:  
(1) \textbf{Pure-speech models}, which handle only speech input and output (e.g., GSLM~\cite{lakhotia2021generative}, AudioLM~\cite{borsos2023audiolm});  
(2) \textbf{Speech-aware models}, which accept both speech and text as input but produce only text output (e.g., Qwen-Audio~\cite{chu2023qwen}, SALMONN~\cite{tang2023salmonn}, Qwen2-Audio~\cite{chu2024qwen2}, WavLLM~\cite{hu2024wavllm}); and  
(3) \textbf{General speech--text models}, which support both modalities at the input and output (e.g., SpeechGPT~\cite{zhang2023speechgpt},  Moshi~\cite{defossez2024moshi} and MiMo-audio~\cite{mimoaudio25}).  
Recent omni models such as Qwen2.5-Omni~\cite{xu2025qwen2} and Qwen3-Omni~\cite{xu2025qwen3} adopt a thinker--talker architecture, which can be viewed as extending a speech-aware LLM to a general speech--text system by integrating a streaming talker (TTS) module.
Motivated by this design, our work focuses on the second category—the speech-aware type—which can be seamlessly upgraded to a general speech--by attaching a streaming TTS system such as CosyVoice~\cite{du2024cosyvoice,du2025cosyvoice}.

While recent advances have pushed speech--LLMs toward broader capabilities, their training often relies heavily on large-scale, manually curated instruction data. \textbf{We argue that extensive task-specific instruction tuning may be unnecessary if the generalization capacity of a frozen LLM can be fully leveraged.} In this paper, we take a step further by introducing both a theoretical and practical framework for aligning the speech modality with LLMs through self-generated data. Specifically, without relying on any instruction-tuning data, we extend a pretrained LLM into a speech--LLM that preserves its original textual understanding ability while achieving robust comprehension of both semantic and paralinguistic aspects of speech.

\textbf{Our main contributions are summarized as follows:}
    \begin{itemize}
    \item We first formulate \textbf{Self-Generated Instruction-Free Tuning (SIFT)} as a training paradigm for speech--LLM models, in which supervision signals are generated by a frozen LLM using \textbf{oracle textual representations of speech} as input, without relying on any explicit instructions.
    \item Building upon the SIFT paradigm, we develop \textbf{AzeroS}, a lightweight speech--LLM trained by updating only projection modules while keeping both the LLM and audio encoders frozen.
    \item Despite minimal training cost and modest data scale, AzeroS achieves state-of-the-art performance on both semantic and paralinguistic benchmarks, including \textit{VoiceBench}~\cite{chen2024voicebench} and \textit{AIR-Bench (Speech)}~\cite{yang2024air}.
\end{itemize}

\section{Related Work}
Training speech--LLMs to achieve general speech understanding remains challenging, as it typically requires large-scale \textbf{Task-Specific Instruction Tuning (TSIT)} data. Common speech-domain instruction tasks include automatic speech recognition (ASR), speech translation (ST), speech emotion recognition (SER), and speech question answering (SQA). 
Qwen-Audio~\cite{chu2023qwen} adopts a multi-task framework that integrates diverse tasks via hierarchical tags, including ASR, ST, SER, language identification (LID), speaker verification (SV), and speaker attribute prediction, etc. 
Qwen2-Audio~\cite{chu2024qwen2} improves this paradigm by using natural-language prompts to unify task instructions. 
SALMONN~\cite{tang2023salmonn} and WavLLM~\cite{hu2024wavllm} employ dual encoders and extensive task-specific instruction tuning. It is found that models trained in this way often become biased toward ASR-style responses probably due to the dominance of ASR data in TSIT distribution. 
To mitigate this, SALMONN applies activation tuning for long-form generative tasks such as storytelling, while WavLLM introduces progressive multi-task mixing to balance single-task overfitting. 
Nevertheless, each task's capability still depends on its corresponding instruction dataset, limiting these models to a fixed set of pre-defined tasks rather than achieving a kind of general open-ended speech understanding.

Given the limitations of TSIT, recent work explores leveraging the inherent generalization ability of LLMs to build instruction-following speech models without task-specific supervision. 
AudioChat-LLaMA~\cite{fathullah2024audiochatllama} aligns speech representations with their transcriptions by ensuring the frozen LLM produces identical responses when fed with either speech or text, thereby enforcing modality consistency. 
BLSP~\cite{wang2023blsp} achieves similar alignment through a continuation-writing objective, allowing speech and text inputs to elicit equivalent LLM behavior. 
Although these methods enable zero-shot generalization across many speech tasks, they primarily focus on aligning semantic information and overlook paralinguistic aspects such as emotion or prosody. 
A follow up work of AudioChat-LLaMA, SpeechEmotionLlama~\cite{kang2024frozen} performs emotion-conditioned response alignment. The resulting system is able to generate more empathetic responses to expressive speech input compared to the baseline~\cite{fathullah2024audiochatllama}. 
Desta2~\cite{lu2024desta2} explores paralinguistic aspects beyond emotion, establishing that a unified description-based training objective exhibits significant generalization capability. The method utilizes a frozen LLM to generate descriptive targets by integrating both speech transcripts and paralinguistic metadata. Zero-shot instruction following ability on both linguistic and paralinguistic aspects is observed, though not using TSIT data. 
Desta2.5~\cite{lu2025desta2} further emphasizes the value of LLM's self-generated data and extends the model's capability beyond speech. To fully utilize the available information in textural description, Desta2.5 crafts thousands of prompts instead of a single descriptive prompt used in Desta2. Despite the advantages,  Desta series still relies on external ASR models for input transcriptions and cannot natively support speech-based interaction.

\section{Methodology}

\subsection{Motivation: Unlocking the Universal Decoder}
\label{sec:motivation}

Our goal is to adapt a pre-trained Large Language Model (LLM) for speech understanding. We posit that modern LLMs are already ``universal decoders''—they possess the inherent ability to follow virtually any instruction if provided with the correct textual information. Therefore, the challenge of multimodal adaptation is not to teach the LLM new tasks, but simply to \textbf{align the audio modality with the text modality} so that the LLM can perceive speech as clearly as it perceives text.

However, the dominant paradigm—Instruction Tuning—often undermines this goal. This method trains the entire Speech-LLM on triplets denoted as $(x, i, y)$, consisting of an audio input $x$, a specific task instruction $i$, and a target response $y$. By optimizing the model to satisfy specific instructions $i$, we introduce two fundamental limitations that hurt generalization:

\begin{enumerate}
    \item \textbf{Instruction-Induced Feature Collapse:} When the system is optimized for a specific instruction $i$ (e.g., ``Transcribe''), the audio encoding layers learn to be ``lazy.'' They extract only the minimal subset of features from $x$ required for that task (e.g., phonemes) and discard the rest (e.g., emotion, speaker identity) as noise. This effectively creates a \textbf{feature filter}, permanently deleting information that might be needed for future, unseen tasks.
    
    \item \textbf{Instruction Overfitting (Modality Neglect):} Strong instruction conditioning encourages the model to ``cheat.'' The MLLM learns spurious correlations between the instruction $i$ and the response $y$ (i.e., modeling $P(y|i)$), effectively bypassing the audio input $x$. The model becomes ``dishonest'' to the sensory input, learning to follow templates rather than grounding its generation in the actual audio content.
\end{enumerate}

To visualize this, consider the \textbf{``Packed Suitcase'' analogy}:
\begin{quote}
    \textit{Imagine the MLLM as a traveler packing for a trip (test time). Instruction tuning is like telling them: ``You are going to the beach.''}
    
    \textit{First, they only pack swimsuits (Feature Collapse). Second, because they are focused on the ``beach'' order, they ignore the weather report (Instruction Overfitting). If they arrive and it is snowing, they fail twice: they lack the right gear, and they didn't even look at the reality before leaving.}
    
    \textit{In contrast, an ideal system should ``pack everything'' and force the traveler to look at the weather. By removing the instruction during training, the model is forced to rely entirely on the audio content to decide what is relevant.}
\end{quote}

We argue that to leverage the LLM's true generalizability, we must abandon task-specific training and instead focus on \textbf{Instruction-Free Input Alignment}—forcing the system to preserve all information in $x$ so the LLM can handle any instruction $i$ at test time.

\subsection{Problem Setup}
\label{sec:setup}
We formally consider a Speech-LLM framework that maps an audio input $x \in \mathcal{X}$ to a textual output $y \in \mathcal{Y}$. The system relies on a frozen LLM, denoted as $g_\phi$, which operates on a continuous embedding space.

The mapping from raw audio to the LLM's input space is defined as $h_\theta(x)$. While $h_\theta$ generally represents the full audio encoding pipeline, in this work we adopt a standard modular architecture where $h_\theta$ is the composition of a \textbf{pre-trained audio encoder} (e.g., Whisper, HuBERT) and a \textbf{learnable projector} (e.g., MLP or Q-Former). For simplicity, \textbf{we freeze the audio encoder and optimize only the projector parameters $\theta$.}

Our objective is to learn a mapping $h_\theta$ that translates acoustic features into the LLM's native embedding space such that the audio becomes ``readable'' by the LLM without altering the LLM itself.

\subsection{Self-Generated Data Construction}
\label{sec:data-construction}
To achieve this alignment without instruction bias, we need an alignment target that represents the ``ideal'' interpretation of the audio. We conceptualize this as a semantic equivalent of the audio's text description.

Let $\tilde{h}(x)$ denote the ``oracle'' textual representation of an audio input $x$ (e.g., the embedding of its ground-truth transcript). To create a supervision signal, we utilize the frozen LLM $g_\phi$ to generate target responses under arbitrary instruction $i$:
\begin{equation}
    y = g_\phi\big(\tilde{h}(x),\, i\big).
\end{equation}

For this process to define a valid alignment target, we pursue two key \textbf{Design Goals}:

\begin{enumerate}
    \item \textbf{Goal 1: Maximizing Information in $\tilde{h}(x)$.} 
    We aim for the oracle text representation $\tilde{h}(x)$ to serve as a comprehensive substitute for the raw audio $x$. Ideally, it captures diverse details—such as content, emotion, and speaker style—to ensure the text embedding represents the audio signal as fully as possible.
    
    \item \textbf{Goal 2: Self-Elicit Capability.} 
    The frozen LLM should \textbf{intrinsically} generate a response $y$ that covers the full information within $\tilde{h}(x)$ without needing an instruction ($i=\varnothing$). 
\end{enumerate}

Based on whether Goal 2 is satisfied, our paradigm branches into two distinct strategies:
\begin{itemize}
    \item \textbf{Method A: SIFT (Optimal).} If the LLM successfully self-elicits the content (Goal 2 Satisfied), we proceed with \textbf{Self-Generated Instruction-Free Tuning (SIFT)}. This is the optimal setting where $i=\varnothing$, forcing pure input alignment.
    \item \textbf{Method B: SIT (Fallback).} If the LLM requires prompting to output the full information (Goal 2 Not Satisfied), we adopt \textbf{Self-Generated Instruction Tuning (SIT)}. For example, we can use a generic prompt \textit{"Describe the audio"} to elicit the target $y$.
\end{itemize}

Under this unified perspective, prior works using similar \textit{self-generated} ideas can be viewed as constrained special cases of our paradigm, as detailed in Table~\ref{tab:comparison}.

\begin{table}[h]
\centering
\caption{\small Unified Comparison of Alignment Strategies. We frame prior works as constrained special cases within our paradigm: AudioChatLlama~\cite{fathullah2024audiochatllama} is limited by information sufficiency (transcript only), while DESTA-2/2.5~\cite{lu2024desta2,lu2025desta2} relies on rigid instructions and auxiliary ASR inputs. SIFT overcomes these bottlenecks by satisfying both Information Maximization and Self-Elicit goals, achieving pure instruction-free alignment.}
\resizebox{\columnwidth}{!}{%
\begin{tabular}{l|c|c|l}
\toprule
\textbf{Method} & \textbf{Oracle Info $\tilde{h}(x)$ (Goal 1)} & \textbf{Instruction $I$ (Goal 2)} & \textbf{Limitations/Characteristics} \\
\midrule
\textbf{AudioChatLlama} & \makecell{Transcript Only \\(e.g., \textit{The weather is great!})} & Null ($\varnothing$) & \makecell[l]{\textbf{Semantic only alignment}, \\ ignores other info.} \\
\midrule
\textbf{DESTA-2/2.5} & \makecell[c]{Transcript + Attributes Tags \\ (e.g., \textit{The weather is great! (Gender: Female, Emotion: Happy...))}} & \makecell[c]{Required \\ (e.g., \textit{"What can you hear from the audio?"})} & \makecell[l]{Still requires instruction $I$; \\ \textbf{Partial alignment,} \\ requires ASR text appended at inference.} \\
\midrule
\textbf{SIT (Fallback)} & 
\multirow{2}{*}{
\begin{tabular}{@{}c@{}}
        \textbf{Full Info}, \\
        either in attributes format \\
        (e.g., \textit{The weather is great! (Gender: Female, Emotion: Happy...)}) \\
        or natural description \\
        (e.g., \textit{A woman says happily:"The weather is great!"...})
    \end{tabular}
    } &\makecell[c]{Required \\ (e.g., \textit{Describe all the information you can hear.})} &\makecell[l]{\textbf{Elicit all info in rich response $y$}; \\ Still relies on instruction $I$, \\may introduce instruction bias.}\\
\cline{1-1} \cline{3-4} 
\textbf{SIFT (Optimal)} & & Null ($\varnothing$) & \makecell[l]{\textbf{Ideal optimal case} \\ \textbf{No instruction bias.} \\ \textbf{Generalized to any task.}} \\
\bottomrule
\end{tabular}%
}
\vspace{0.2cm}
\label{tab:comparison}
\end{table}

\subsection{SIFT: Self-Generated Instruction-Free Tuning}
\label{sec:sift-training}
We primarily focus on the optimal setting: SIFT. Our objective is \textbf{Input-Level Audio-Text Alignment}: we aim to train the encoder $h_\theta$ such that the projected audio features become indistinguishable from the oracle text embeddings from the perspective of the LLM:
\begin{equation}
    h_\theta(x) \approx \tilde{h}(x).
\end{equation}

We achieve this by minimizing the prediction error on the self-generated target $y$ \textbf{without} providing any instruction ($i=\varnothing$). The optimization problem is:
\begin{equation}
    \theta^* = \operatorname*{argmin}_\theta \mathcal{L}_{\text{CE}}\Big( y,\, g_\phi\big(h_\theta(x), \varnothing\big) \Big).
\end{equation}
By removing the instruction variable, we force the gradient to flow entirely through the encoder, driving it to bridge the modality gap.

\subsection{Why Instruction-Free Generalizes Best}
\label{sec:intuition}
The advantage of SIFT is best understood through the lens of alignment completeness.

\textbf{1. Instruction Tuning yields Partial Alignment:}
When training with a specific instruction $I$ (maximizing $P(Y \mid X, I)$), the model only needs to align the subset of features in $X$ relevant to $I$ (e.g., phonemes for ASR), often ignoring paralinguistic details. This results in a \textbf{task-dependent alignment} that fails when the LLM is asked to perform a new task requiring different features.

\textbf{2. SIFT yields General Input Alignment:}
By removing the instruction (maximizing $P(Y \mid X)$), we remove the filter. Because the target $y$ reflects the full content of the oracle text $\tilde{h}(x)$, the only way to minimize the loss is for the encoder to achieve \textbf{general alignment}. The projected audio must capture all auditory details to reproduce the unguided textual output.

Mathematically, blocking the instruction channel forces the encoder to solve the hardest alignment problem:
\begin{equation}
    h_\theta(x) \xrightarrow{\text{align}} \tilde{h}(x) \quad \text{(Globally)}.
\end{equation}
This produces a ``universal'' encoder that maps audio into the LLM's space faithfully. At test time, we can simply plug in any instruction $I_{new}$, and the frozen LLM—being a universal decoder—will execute it successfully because the input is perfectly aligned.

\section{AZeroS: System Overview}
To verify the effectiveness of the proposed SIFT paradigm, as illustrated in Figure~\ref{fig:overview}, we build a speech--LLM named AzeroS (\textbf{A}uden \textbf{Zero}-instruction-tuned \textbf{S}peech-LLM), implemented within the open-source Auden toolkit\footnote{\url{https://github.com/AudenAI/Auden}}.

\subsection{Model Architecture}
AzeroS adopts a typical encoder–projector–LLM architecture, following the design of most multimodal LLMs.

\textbf{Large Language Model.} We select Qwen2.5-7B-Instruct~\cite{yang2024qwen2} as our backbone LLM. Qwen2.5 is a high-performance open-source language model series trained on large-scale data, and its 7B Instruct variant provides strong instruction-following capability and consistently competitive performance across diverse benchmarks.

\textbf{Semantic Encoder.} We adopt the TTA encoder~\cite{auden_tta}, a 153M Zipformer-based~\cite{yao2023zipformer} model trained on 358k hours of multilingual ASR and translation data, to obtain high-quality semantic representations. The encoder operates on 80-dimensional log-Mel filterbank features computed with a 25 ms window and a 10 ms stride, and produces 768-dimensional feature sequences at a frame rate of 25 Hz after a 4× temporal downsampling.

\textbf{Paralinguistic Encoder.} To capture speaker-specific and paralinguistic cues (e.g., emotion, prosody, speaker identity), we adopt the Auden-Voice encoder~\cite{auden_voice}, a 153M Zipformer-based model trained on 8k hours of multi-task data covering speaker identity, age, gender, and emotion. The encoder produces 768-dimensional paralinguistic embeddings at a frame rate of 25 Hz.

\textbf{Projector.} Each speech encoder is paired with a linear 23.8M projector that downsamples and maps the encoder features into the LLM embedding space. The projector groups every 4 frames and applies a two-layer MLP with a ReLU activation. This reduces the frame rate to 6.25 Hz, aligning the audio representations with the temporal granularity of text tokens.

\subsection{Training Paradigm}
Following the self-generated paradigm introduced in Section~\ref{sec:sift-training}, AzeroS integrates data generation and model training into a paired process.

\textbf{Data Generation.}
Given a speech--text pair $(x, \tilde{h}(x))$, where $x$ denotes the speech signal and $\tilde{h}(x)$ is the textual representation we aim to map (e.g., its original supervision such as transcript or metadata), we use $\tilde{h}(x)$ as the sole input to the LLM to generate a supervision response $y$. This produces a training sample of the form $(x, y)$. 


\textbf{Model Training.}
During training, both the LLM and the audio encoders remain frozen. Only the lightweight projector (23.8M parameters) is updated, allowing it to focus solely on learning the desired mapping from audio representations into the LLM embedding space.

\textbf{Multi-stage Alignment.}
To better align both the semantic and paralinguistic information in speech with the LLM, we divide training into two stages. In the first stage, only semantic information is considered: we use the transcript as $\tilde{h}(x)$ and align the TTA encoder alone. In the second stage, $\tilde{h}(x)$ is extended to include both the transcript and the associated meta information, enabling the model to learn paralinguistic mappings as well. Since the semantic component has already been aligned in Stage~1, we freeze all modules except the paralinguistic projector and update only this component during Stage~2.

\section{Experimental Setup and Results}
\subsection{Data Source}

\begin{wraptable}{r}{0.5\textwidth}
\vspace{-0.5cm}
\centering
\small
\caption{\small Summary of data source used for semantic and paralinguistic alignment in AZeroS.}
\label{tab:data_mixing}

\adjustbox{max width=0.5\textwidth}{
\begin{tabular}{l c c}
\toprule
\textbf{Dataset} & \textbf{Label} & \textbf{Hours (h)} \\
\midrule
\rowcolor{secblue}
\multicolumn{3}{l}{\textbf{Data-SP}} \\
IEMOCAP~\cite{busso2008iemocap}      & gender, emotion, transcript$^1$       & 9  \\
CREMA-D~\cite{cao2014crema}        & gender, age, emotion, transcript$^1$  & 5  \\
MELD~\cite{poria2019meld}         & gender, emotion, transcript$^1$     & 8  \\
RAVDESS~\cite{livingstone2018ryerson}      & gender, emotion, transcript$^1$       & 1  \\
TESS~\cite{SP2/E8H2MF_2020}         & gender, age, emotion, transcript$^1$  & 1  \\
DailyTalk~\cite{lee2023dailytalk}   & emotion, transcript               & 21 \\
CommonVoice-en~\cite{ardila2019common}  & gender, age, transcript            & 1199 \\
AISHELL-1~\cite{bu2017aishell}      & gender, transcript                 & 150 \\
EmotionTalk~\cite{sun2025emotiontalk}       & gender, emotion, transcript        & 23 \\
CS-Dialogue~\cite{zhou2025cs}           & gender, age, transcript           & 104 \\
VoxCeleb2~\cite{chung2018voxceleb2}    & gender, transcript$^1$               & 2026 \\
\rowcolor{secblue}
\textbf{Total} &  & \textbf{3548$^{2}$} \\
\midrule
\rowcolor{secblue}
\multicolumn{3}{l}{\textbf{Data-S}} \\
WenetSpeech~\cite{zhang2022wenetspeech}  & transcript & 9685 \\
GigaSpeech~\cite{chen2021gigaspeech}   & transcript & 9396 \\
CommonVoice~\cite{ardila2019common}  & transcript & 3302 \\
\rowcolor{secblue}
\textbf{Total} & & \textbf{22383} \\
\bottomrule
\end{tabular}
}
\vspace{-4mm}
\begin{flushleft}
\scriptsize
\item[1] Transcripts are generated by the TTA~\cite{auden_tta} model. \\
\item[2] Datasets are rebalanced during training.
\end{flushleft}
\vspace{-1cm}
\end{wraptable}

Table~\ref{tab:data_mixing} summarizes all datasets used in our experiments. All datasets are publicly available.
Based on the available supervision types, we categorize the data into two sets:

\textbf{Data-SP.} This set comprises 3{,}548 hours of speech with both semantic transcripts and paralinguistic annotations. For corpora that contain paralinguistic annotations but lack transcripts, we generate transcripts using the TTA ASR model~\cite{auden_tta}.

\textbf{Data-S.} This set consists of two standard Chinese and English ASR corpora together with the multilingual Common Voice corpus. It contains 22{,}383 hours of speech in total and provides semantic supervision in the form of ASR transcripts only.

\subsection{Evaluation Benchmarks}
\textbf{Voicebench}~\cite{chen2024voicebench}. We use VoiceBench to evaluate the semantic understanding ability of AZeroS. VoiceBench is specifically designed to assess speech--LLMs in realistic spoken-instruction scenarios, covering diverse tasks such as general knowledge, reasoning, and safety across both real and synthetic speech. Unlike single-task evaluations such as ASR or speech translation, its diversity enables a more comprehensive assessment of a model’s ability to generalize to \textbf{unseen spoken tasks} and instructions within semantic understanding. We follow the official evaluation protocol\footnote{gpt-4o-2025-01-01-preview is used for evaluation.} and report instruction-following and comprehension accuracy.

\textbf{AIR-Bench-Speech}~\cite{yang2024air}.
To evaluate paralinguistic understanding and complement the semantic evaluation, we adopt the speech-specific subset of AIR-Bench.
AIR-Bench encompasses both semantic and non-semantic aspects of speech; our selection focuses on paralinguistic cues such as prosody, emotion, and speaker characteristics, while still retaining semantic components in speech-only tasks.
The \emph{Foundation} part contains multiple-choice questions\footnote{An additional prompt is added to ensure the models only output the option.}, and the \emph{Chat}\footnote{gpt-4-turbo-2025-01-01-preview is used for evaluation} part consists of open-ended question–answer pairs, both formatted with text-based instructions appended after the audio input.
This setup provides a compact yet comprehensive evaluation of how well the model integrates non-linguistic cues with semantic information in spoken input.

\subsection{Stage 1: Semantic Alignment} \label{sec:stage_1}
\begin{figure*}[h]
  \centering
  \includegraphics[width=0.9\linewidth]{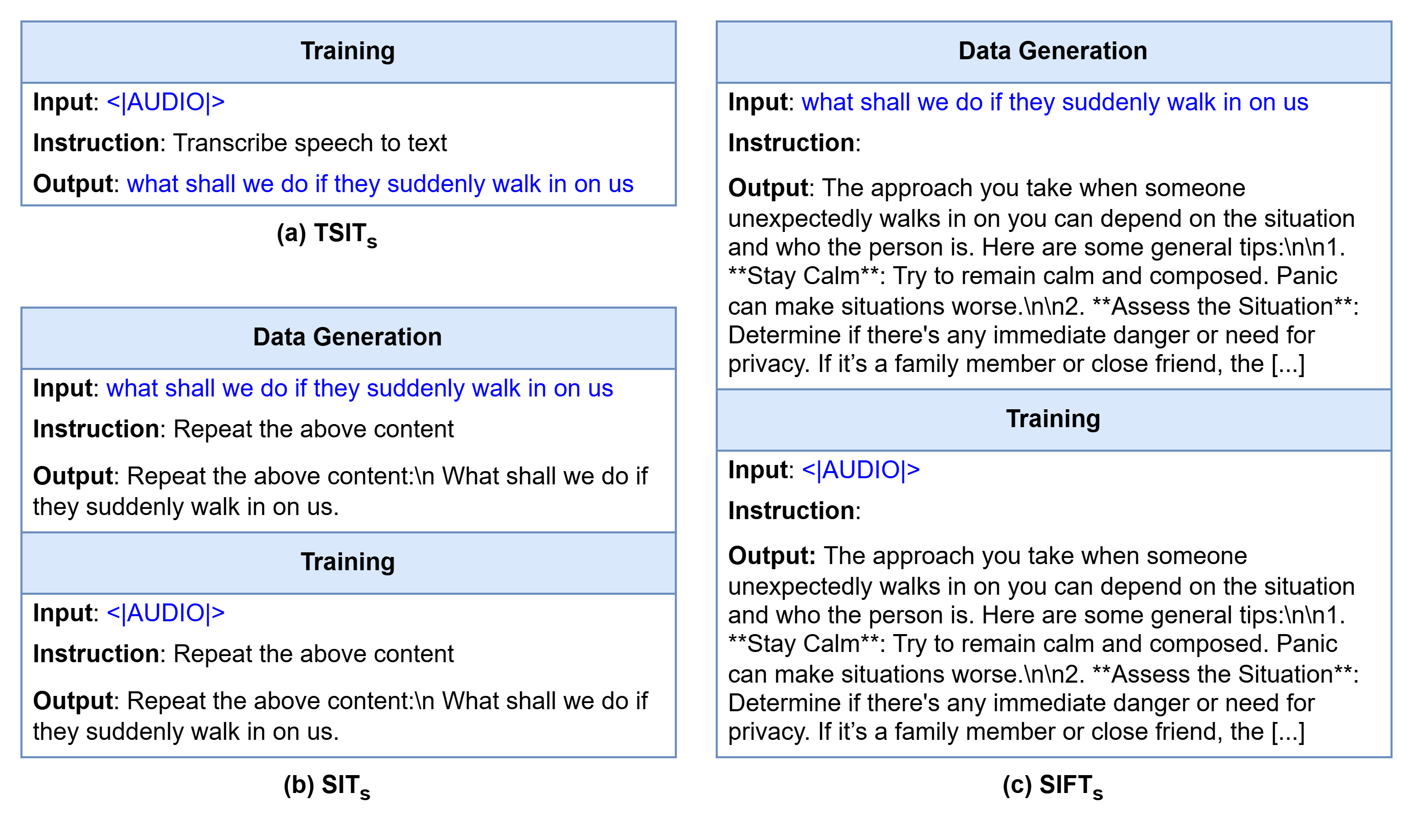}
  \vspace{-0.5cm}
  \caption{
\small
Illustration of semantic-only instruction-tuning configurations:
(a) \textbf{TSIT\textsubscript{s}}, (b) \textbf{SIT\textsubscript{s}}, and (c) \textbf{SIFT\textsubscript{s}}.
Examples are selected from an utterance in the IEMOCAP dataset~\cite{busso2008iemocap},
with the transcript \textcolor{blue}{\textit{``what shall we do if they suddenly walk in on us''}}.
(\truncated indicates truncated text. Detailed examples are in Appendix~\ref{app:example}.)
}
  \label{fig:sift_s}
\end{figure*}

\textbf{Experimental Setup.} We first focus on semantic information alignment. Accordingly, the model is configured to include only the TTA encoder and its associated projector. We start with preliminary experiments on the small scale Data-SP set and compare three instruction-tuning configurations (the subscript $s$ denotes \textit{semantic}; see Figure~\ref{fig:sift_s} for examples):
\begin{itemize}
    \item (a) \textbf{Task-Specific Instruction Tuning (\textbf{TSIT\textsubscript{s}})}: The transcript is used directly as the supervision target, while the instruction is randomly sampled from a predefined set of ASR-specific prompts.
\item (b) \textbf{Self-Generated Instruction Tuning (\textbf{SIT\textsubscript{s}})}: The transcript is provided as input to the LLM together with an additional instruction that guides the LLM to generate an ASR-style response.
\item (c) \textbf{Self-Generated Instruction-Free Tuning (\textbf{SIFT\textsubscript{s}})}: The transcript is used as the sole input without any accompanying instruction, and the supervision $y$ is obtained from the LLM’s open-ended response.
\end{itemize}

\begin{table*}[t]
\caption{\small
Preliminary results on \textit{VoiceBench} comparing different instruction-tuning configurations (see Figure~\ref{fig:sift_s}) for \textbf{semantic-only} alignment, trained on the \textbf{Data-SP} subset.
}
\centering
\renewcommand{\arraystretch}{1}
\setlength{\tabcolsep}{3.8pt}
\small
\begin{adjustbox}{max width=\textwidth}
\begin{threeparttable}
\begin{tabular}{lccccccccc
>{\columncolor{secblue}}c}
\toprule
\textbf{Method} &
\textbf{AlpacaEval} &
\textbf{CommonEval} &
\textbf{WildVoice} &
\textbf{SD-QA} &
\textbf{BBH} &
\textbf{AdvBench} &
\textbf{IFEval} &
\textbf{OBQA} &
\textbf{MMSU} &
\textbf{Overall} \\
& (GPT) & (GPT) & (GPT) & (Panda/GPT) & (Acc.) & (Refusal) & (P./I. Acc.) & (Acc.) & (Acc.) & \\
\midrule
TSIT\textsubscript{s}  & 1.12 & 1.34 & 1.10 & 4.52 & 47.80 & 3.46 & 24.13 & 24.62 & 24.59 & 22.26 \\
SIT\textsubscript{s}  & 3.98 & 3.74 & 3.38 & 45.75 & 51.00 & 95.96 & 39.09 & 34.07 & 33.12 & 57.89 \\
SIFT\textsubscript{s} & 4.32 & 4.14 & 3.74 & 55.70 & 55.60 & 98.27 & 54.40 & 59.34 & 51.01 & \textbf{68.70} \\
\bottomrule
\end{tabular}
\end{threeparttable}
\end{adjustbox}
\label{tab:s_results}
\end{table*}

\textbf{Results and Analysis.} Preliminary experimental results on VoiceBench are shown in Table~\ref{tab:s_results}. In \textbf{TSIT\textsubscript{s}}, the speech--LLM is optimized purely for the ASR objective. Consistent with prior observations~\citep{tang2023salmonn, hu2024wavllm, wang2023blsp}, such training encourages the model to memorize ASR-specific behaviors and hampers its ability to generalize to unseen instructions at inference time, resulting in a poor overall score of 22.26. These results further confirm that directly aligning speech and text through raw ASR supervision is insufficient for learning a generalizable speech--text interface.

By contrast, the self-generated variant \textbf{SIT\textsubscript{s}}, while still grounded in the same ASR task, achieves substantially improved generalization across all VoiceBench tasks, reaching an overall score of 57.89. We observe that, compared to the original transcripts, self-generated supervision often introduces auxiliary phrasing, subtle linguistic variations, or occasional follow-up questions. Such variations impose stricter constraints on the output space, which in turn encourage more robust alignment on the input side and mitigate overfitting to rigid ASR patterns.

Finally, \textbf{SIFT\textsubscript{s}} achieves the best overall performance, with a score of 68.70, indicating the most generalizable semantic alignment. This improvement can be attributed to its instruction-free formulation, which removes explicit task cues during training and enables the model to develop a more general semantic understanding applicable to diverse voice-based queries. \textbf{Based on these results, we adopt \textbf{SIFT\textsubscript{s}} as the default and most effective configuration for Stage~1 semantic alignment.}

\subsection{Stage 2: Joint Semantic and Paralinguistic Alignment}
\label{sec:stage_2}
\textbf{Experimental Setup.} To extend AzeroS to jointly understand semantic and paralinguistic information, we proceed to a second training stage in which both types of information are aligned.
As discussed in Section~\ref{sec:stage_1}, \textbf{SIFT\textsubscript{s}} achieves the best performance in semantic understanding. We therefore scale up the Stage~1 training from 3k-hour \textbf{Data-SP} to the 22k-hour \textbf{Data-S} set to obtain stronger semantic alignment baseline. After this stage, the TTA encoder and its corresponding projector are kept frozen. In Stage~2, only the paralinguistic projector on top of the Voice Encoder is trained and only \textbf{Data-SP} with paralinguistic labels are used.

\begin{figure*}[h]
  \centering
  \includegraphics[width=0.9\linewidth]{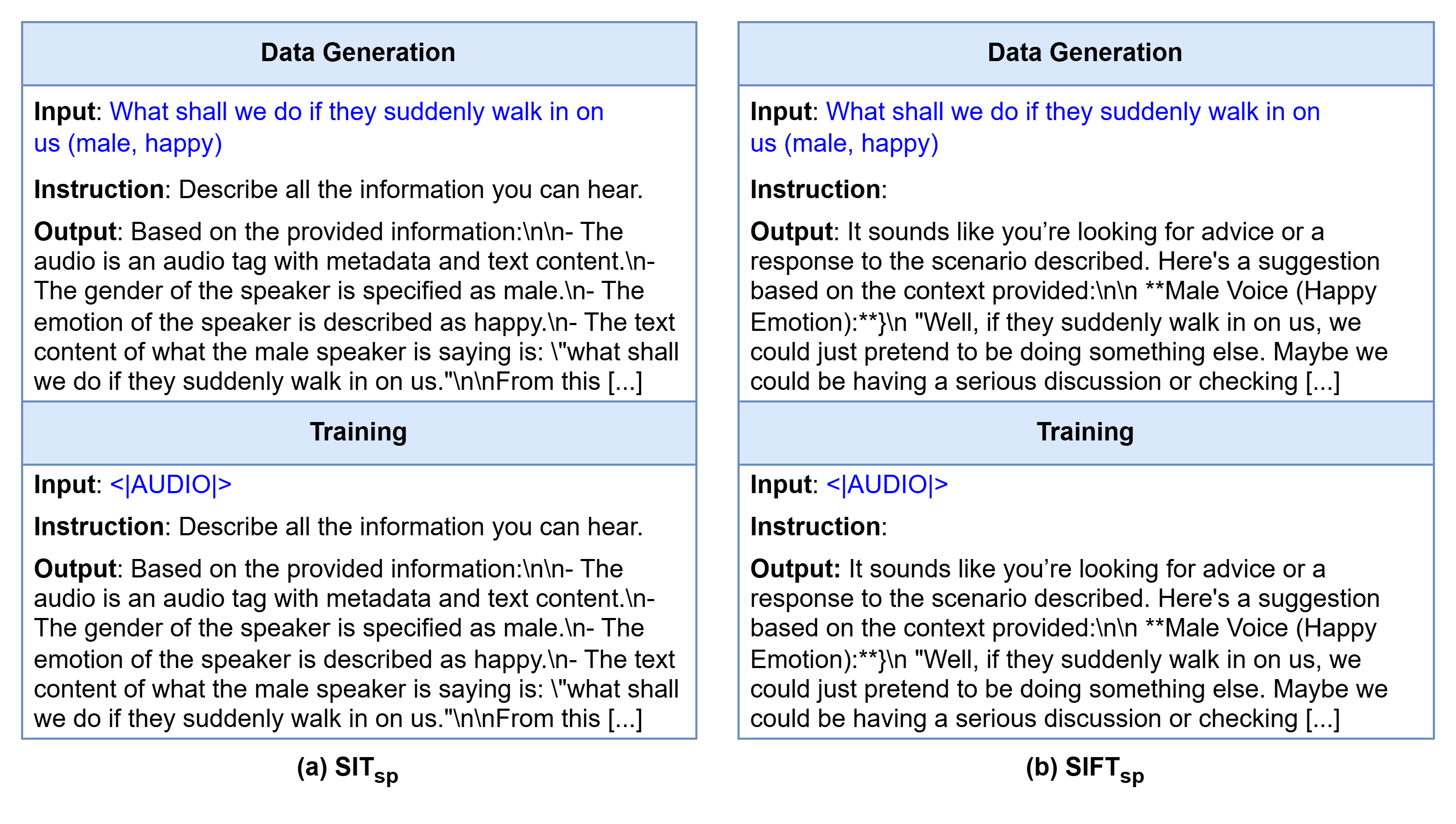}
  \vspace{-0.5cm}
  \caption{
\small
Illustration of semantic and paralinguistic tuning configurations:
(a) \textbf{SIT\textsubscript{sp}} and (b) \textbf{SIFT\textsubscript{sp}}.
Examples are selected from an utterance in the IEMOCAP dataset~\cite{busso2008iemocap},
with the transcript \textcolor{blue}{\textit{"what shall we do if they suddenly walk in on us"}} and paralinguistic metadata \textcolor{blue}{\textit{(gender: female, emotion: happy)}}.
(\truncated indicates truncated text. Detailed examples are in Appendix~\ref{app:example}.)
}
  \label{fig:sift_sp}
\end{figure*}

We compare two tuning configurations for joint semantic and paralinguistic alignment (the subscript \(sp\) denotes \textit{semantic + paralinguistic}; see Figure~\ref{fig:sift_sp} for examples):
\begin{itemize}
    \item (a) \textbf{Self-Generated Instruction Tuning (\textbf{SIT\textsubscript{sp}})}: Both the transcript and paralinguistic metadata (e.g., gender, age, and emotion) are provided as input to the LLM under a generic instruction \textit{"Describe all the information you can hear."}, and the generated response is used as training supervision.
    \item (b) \textbf{Self-Generated Instruction-Free Tuning (\textbf{SIFT\textsubscript{sp}})}: This configuration is identical to SIT\textsubscript{sp} except that no instruction is provided.
\end{itemize}

\begin{table*}[htbp!]
\scriptsize
\resizebox{\textwidth}{!}{
\begin{threeparttable}
\caption{
\small Full results on VoiceBench and AIR-Bench, comparing multi-stage training configurations for comprehensive \textbf{semantic + paralinguistic alignment}, trained on both \textbf{Data-S} and \textbf{Data-SP}.
}
\centering
\setlength{\tabcolsep}{1pt}
\renewcommand{\arraystretch}{1.1}
\begin{tabular}{
  l
  *{9}{>{\centering\arraybackslash}p{0.77cm}}
  >{\columncolor{secblue}\centering\arraybackslash}p{0.75cm} %
  *{6}{>{\centering\arraybackslash}p{0.75cm}}
  >{\columncolor{secblue}\centering\arraybackslash}p{0.75cm}
  >{\columncolor{secblue}\centering\arraybackslash}p{0.75cm}
}
\toprule
& \multicolumn{10}{c}{\textbf{VoiceBench}}  
& \multicolumn{8}{c}{\textbf{AIR-Bench (Speech)}} \\
\cmidrule(lr){2-11}  \cmidrule(lr){12-19} 
\textbf{Method} 
& \makecell{Alpaca\\Eval} & \makecell{Comm\\Eval} & \makecell{Wild\\Voice} & \makecell{SD-\\QA}
& BBH & \makecell{Adv\\Bench} & \makecell{IF\\Eval} & OBQA & MMSU & Overall
& Gender & Emotion & Age & LID & Entity & Intent & Avg & Chat \\
\midrule
&\multicolumn{18}{c}{\textbf{Stage 1: semantic-only (on Data-S)}} \\
\midrule
SIFT\textsubscript{s}   &4.44 	&4.22 	&3.95 	&59.22 	&56.10 	&99.23 	&62.63 	&73.19 	&60.08 	&\textbf{73.63}  &30.47 	&46.40 	&34.60 	&76.20 	&72.40 	&81.80 	&56.98 	&7.22  \\
\midrule
&\multicolumn{18}{c}{\textbf{Stage 2: semantic + paralinguistic (on Data-SP)}} \\
\midrule
SIT\textsubscript{sp}   & 4.38 	& 4.14  & 3.86  & 57.60 & 55.40 & 98.65 & 60.01 & 71.87 & 58.85 & 72.22 & {74.85} & 73.10 & {57.20} & 85.30 & 72.90 & 87.20	& 75.09 & 8.19 \\
SIFT\textsubscript{sp}  & 4.41  & 4.08  & 3.81  & 59.13 & 55.80 & 99.04 & 61.10 & 70.33 &59.01  & 72.27 &83.18  & 74.60 & 64.00 & 85.50 & 72.70 & 86.10 & \textbf{77.68} & 8.21 \\
\textbf{SIFT\textsubscript{s} + SIFT\textsubscript{sp}}  & 4.44  & 4.18  & 3.91  & 60.22 & 56.30 & 98.65 & 61.29 & 72.09 & 59.01 & \textbf{73.13} & 86.75 & 71.45 & 61.30 & 84.80 & 73.60 & 85.60 & 77.25 & \textbf{8.28} \\
\bottomrule
\end{tabular}
\label{tab:final_results}
\end{threeparttable}
}
\end{table*}

\textbf{Results and Analysis.}
Results on both VoiceBench and AIR-Bench are summarized in Table~\ref{tab:final_results}. Comparing the \textbf{SIFT\textsubscript{s}} results in Table~\ref{tab:final_results} with those in Table~\ref{tab:s_results}, training on the larger \textbf{Data-S} corpus leads to consistently better semantic understanding than training on \textbf{Data-SP} (e.g., 73.63 vs.\ 68.70 overall on VoiceBench). However, without explicit paralinguistic alignment, the model performs poorly on AIR-Bench, particularly on non-semantic tasks such as gender, age, and emotion recognition.

In Stage~2, jointly aligning semantic and paralinguistic information using Data-SP substantially improves performance on AIR-Bench for both \textbf{SIT\textsubscript{sp}} and \textbf{SIFT\textsubscript{sp}}. Consistent with the observations in Stage~1, the instruction-free variant \textbf{SIFT\textsubscript{sp}} continues to outperform \textbf{SIT\textsubscript{sp}}, although the performance gap is smaller. This can be attributed to the nature of AIR-Bench, which primarily evaluates in-domain, seen tasks, limiting the advantage of instruction-free training in terms of generalization.

We further observe a slight degradation in VoiceBench performance after Stage~2, despite freezing the TTA branch. To alleviate this issue, we combine both \textbf{SIFT\textsubscript{s}} and \textbf{SIFT\textsubscript{sp}} on Data-SP during training in Stage~2, yielding the best overall performance and is thus selected as our final model. This result also suggests that incorporating more diverse training signals—and potentially more data—could further improve the model in future work.

\begin{table*}[t]
\scriptsize
\caption{
\small
Comparison of AzeroS with text-only, cascaded, and end-to-end speech–LLM systems on VoiceBench and AIR-Bench.  
Qwen2.5 refers to the Qwen2.5-7B-Instruct.  
AzeroS nearly reaches the text-only semantic upper bound on VoiceBench while substantially outperforming all models on AIR-Bench, achieving the best overall balance between semantic and paralinguistic understanding. Closed-source systems are shown in \textcolor{gray}{gray} for clarity.
}
\centering
\setlength{\tabcolsep}{1pt}
\renewcommand{\arraystretch}{1.1}
\begin{adjustbox}{max width=\textwidth}
\begin{threeparttable}
\begin{tabular}{
  l
  *{9}{>{\centering\arraybackslash}p{0.77cm}}
  >{\columncolor{secblue}\centering\arraybackslash}p{0.75cm} %
  *{6}{>{\centering\arraybackslash}p{0.75cm}}
  >{\columncolor{secblue}\centering\arraybackslash}p{0.75cm}
  >{\columncolor{secblue}\centering\arraybackslash}p{0.75cm}
}
\toprule
& \multicolumn{10}{c}{\textbf{VoiceBench}}  
& \multicolumn{8}{c}{\textbf{AIR-Bench (Speech)}} \\
\cmidrule(lr){2-11}  \cmidrule(lr){12-19} 
\textbf{Model} 
& \makecell{Alpaca\\Eval} & \makecell{Comm\\Eval} & \makecell{Wild\\Voice} & \makecell{SD-\\QA}
& BBH & \makecell{Adv\\Bench} & \makecell{IF\\Eval} & OBQA & MMSU & Overall
& Gender & Emotion & Age & LID & Entity & Intent & Avg & Chat \\
\midrule
\multicolumn{18}{l}{\textit{\textbf{Text Only Model}}} \\
Qwen2.5 &4.66 &4.55 &4.62 	&62.03 	&80.00 	&99.04 	&70.14 	&84.84 &71.57 &82.69 & N/A & N/A & N/A & N/A & N/A & N/A & N/A & N/A\\ 
Qwen2.5 (TN) &4.61 	&4.53 	&4.56 	&63.84 	&56.30 	&98.85 	&66.11 	&74.07 &64.51 &77.52 & N/A & N/A & N/A & N/A & N/A & N/A & N/A & N/A\\
\midrule
\multicolumn{18}{l}{\textit{\textbf{Cascaded System}}} \\
\textcolor{gray}{Whisper+GPT-4o} & \textcolor{gray}{4.80} & \textcolor{gray}{4.47} & \textcolor{gray}{4.62} & \textcolor{gray}{75.77} & \textcolor{gray}{87.20} & \textcolor{gray}{98.27} & \textcolor{gray}{76.51} & \textcolor{gray}{92.97} & \textcolor{gray}{81.69} & \textcolor{gray}{87.80} & \textcolor{gray}{21.90} & \textcolor{gray}{59.50} & \textcolor{gray}{41.10} & \textcolor{gray}{96.80} & \textcolor{gray}{69.80} & \textcolor{gray}{87.70} & \textcolor{gray}{62.80} & \textcolor{gray}{7.54} \\

Whisper+Qwen2.5 &4.64 &4.33 &4.21 &58.50  &52.85 	&98.27 	&63.99 	&78.24 	&69.00 	&76.05  &28.36 	&50.80 	&36.40 	&88.00 	&73.60 	&82.70 	&59.98 	&7.34\\
\midrule
\multicolumn{18}{l}{\textit{\textbf{End-to-end Speech-LLM}}} \\
\textcolor{gray}{GPT-4o} &\textcolor{gray}{4.78} & \textcolor{gray}{4.49} & \textcolor{gray}{4.58} & \textcolor{gray}{75.50} & \textcolor{gray}{84.10} & \textcolor{gray}{98.65} & \textcolor{gray}{76.02} & \textcolor{gray}{89.23} & \textcolor{gray}{80.25} & \textcolor{gray}{86.75} & \textcolor{gray}{*} &\textcolor{gray}{49.10} &\textcolor{gray}{*} &\textcolor{gray}{76.00} &\textcolor{gray}{61.60} & \textcolor{gray}{85.8} & \textcolor{gray}{*} & \textcolor{gray}{7.53}\\
\textcolor{gray}{Gemini2.5-pro}& -& - & - & - & - & - & - &- &-&- &\textcolor{gray}{90.7} & \textcolor{gray}{60.70} & \textcolor{gray}{34.10} & \textcolor{gray}{99.10} & \textcolor{gray}{68.5} & \textcolor{gray}{92.2} & \textcolor{gray}{74.22} & \textcolor{gray}{8.52} \\
Moshi &2.01 &1.60 &1.30 &15.64 	&47.40 	&44.23 	&10.12 	&25.93 	&24.04 	&29.51 & N/A & N/A & N/A & N/A & N/A & N/A & N/A & N/A \\ 
SALMONN &-&-&-&-&-&-&-&-&-&-&35.5 & 29.9 & 48.7 & 28.1 &51.7 & 36.7 &38.43 &6.16\\
Phi-4-multimodal&3.81 &3.82 &3.56 &39.78 &61.80  &100.00 &45.35 &65.93 	&42.19 	&64.32 &- & -& - & - & - & - & - & - \\
GLM-4-Voice	&3.97 	&3.42 	&3.18 	&36.98 	&52.80 	&88.08 	&25.92 	&53.41 	&39.75 	&56.48 & 23.91 & 22.95 & 18.70 & 25.40 & 27.90 & 21.10 & 23.33 & 5.53 \\
Qwen2-Audio	&3.42 	&3.29 	&2.76 	&31.65 	&53.00 	&99.04 	&26.35 	&48.35 	&36.14 	&53.77 	&64.71 	&48.15 	&23.10 	&77.80 	&87.00 	&84.70 	&64.24 	&7.20  \\
DeSTA2.5    &3.73 &2.52 &3.30 &46.47 &62.40 &97.69 &65.47 &72.75 &58.56 &66.04 &84.24 &64.30 &65.60 &97.30 &65.20 &83.70 &76.72 &7.57 \\
Qwen2.5-Omni	&3.88 	&3.77 	&3.52 	&46.75 	&63.70 	&97.31 	&40.19 	&81.54 &61.45 &68.26 &89.76 	&54.85 	&44.80 	&89.70 	&79.70 	&88.60 	&74.57 	&6.97 \\
Qwen3-Omni-30B & 4.74 & 4.54 & 4.58 & 76.90 & 80.40 & 99.30 & 77.80 & 89.70 & 68.10 & \underline{85.49} & 91.11 & 62.20 & 36.90 & 97.70 & 80.40 & 90.70 & 76.50 & 7.85 \\
\textbf{AzeroS (ours)}  & 4.44  & 4.18  & 3.91  & 60.22 & 56.30 & 98.65 & 61.29 & 72.09 & 59.01 & \textbf{73.13} & 86.75 & 71.45 & 61.30 & 84.80 & 73.60 & 85.60 & \textbf{77.25} & \textbf{8.28} \\
\bottomrule
\end{tabular}
\begin{flushleft}
\scriptsize
* reject by content management policy. \\
\end{flushleft}
\end{threeparttable}
\label{tab:compare}
\end{adjustbox}
\end{table*}

\subsection{Comparison to Other Systems}
Table~\ref{tab:compare} compares AzeroS with representative text-only, cascaded, and end-to-end speech–LLM systems.

\textbf{Text-only Model.}
Since our self-generated instruction tuning paradigm freezes the LLM and assumes perfect reasoning given textual input, a text-only model serves as the theoretical upper bound.
We evaluate Qwen2.5-7B-Instruct on VoiceBench using the original text of spoken instructions. 
The model is also observed to be highly sensitive to special punctuation (e.g., quotation marks, colons), so we additionally test a text-normalized (TN) version that removes them. 
This simple preprocessing yields \textbf{an overall score of 77.52, providing an approximate semantic ceiling of our LLM backbone.} 
As it does not process audio, this model is not evaluated on AIR-Bench.

\textbf{Cascaded Systems.}
Cascaded pipelines combine an ASR front-end and a textual LLM back-end.
Using Whisper-large-v3~\cite{radford2023robust} as the ASR module, we test both Qwen2.5 and GPT-4o~\cite{hurst2024gpt} as LLMs.
Their large gap on VoiceBench (76.05 vs.\ 87.80) reflects differences in reasoning ability rather than ASR accuracy.
However, both perform poorly on AIR-Bench, as they lack access to paralinguistic cues such as emotion, tone, and style.

\textbf{End-to-end Speech-LLM.}
End-to-end speech–LLMs directly map audio to textual responses and are often expected to mitigate cascading errors introduced by ASR. 
However, as shown in Table~\ref{tab:compare}, most existing end-to-end models still significantly underperform cascaded systems on VoiceBench, indicating weaker semantic alignment. 
We attribute this gap not to the end-to-end formulation itself, but to the limited generalization of conventional task-specific instruction-tuning strategies commonly used in their training. 
In contrast, end-to-end models excel on AIR-Bench, where their direct access to paralinguistic cues (e.g., emotion, tone, and speaking style) provides an advantage. 
Compared with these systems, AzeroS—trained with our self-generated instruction-tuning framework—achieves a superior balance between semantic and paralinguistic understanding. 
Within the Qwen2.5 family, AzeroS reaches 72.91 on VoiceBench, approaching the text-only (77.52) and cascaded (76.05) upper bounds, while substantially outperforming other end-to-end variants on AIR-Bench.
We also observe a significant performance gain in Qwen3-Omni-30B, particularly on semantic tasks. We attribute this improvement to its stronger and larger backbone, suggesting that \textbf{applying our proposed training framework to more capable base models will yield further gains.} This highlights the scalability and potential of our method as pretrained LLM backbones continue to advance.

\subsection{Ablation Study}

\begin{table*}[htbp!]
\scriptsize
\centering
\caption{
\small
Ablation results on VoiceBench and AIR-Bench, evaluating the impact of model architecture and training strategy. 
The 1-stage setting merges all training data and performs joint training in a single stage, in contrast to the proposed two-stage strategy.
}

\setlength{\tabcolsep}{1pt} 
\renewcommand{\arraystretch}{1} 
\begin{adjustbox}{max width=\textwidth}
\begin{threeparttable}
\begin{tabular}{
  l
  *{9}{>{\centering\arraybackslash}p{0.77cm}}
  >{\columncolor{secblue}\centering\arraybackslash}p{0.75cm} 
  *{6}{>{\centering\arraybackslash}p{0.75cm}}
  >{\columncolor{secblue}\centering\arraybackslash}p{0.75cm} 
  >{\columncolor{secblue}\centering\arraybackslash}p{0.75cm} 
}
\toprule
& \multicolumn{10}{c}{\textbf{VoiceBench}}  
& \multicolumn{8}{c}{\textbf{AIR-Bench (Speech)}} \\
\cmidrule(lr){2-11}  \cmidrule(lr){12-19} 
\textbf{System} 
& \makecell{Alpaca\\Eval} & \makecell{Comm\\Eval} & \makecell{Wild\\Voice} & \makecell{SD-\\QA}
& BBH & \makecell{Adv\\Bench} & \makecell{IF\\Eval} & OBQA & MMSU & Overall
& Gender & Emotion & Age & LID & Entity & Intent & Avg & Chat \\
\midrule
\multicolumn{19}{l}{\textit{\textbf{2 Stages (AZeroS)}}} \\
TTA-Voice
 & 4.44  & 4.18  & 3.91  & 60.22 & 56.30 & 98.65 & 61.29 & 72.09 & 59.01 & \textbf{73.13} & 86.75 & 71.45 & 61.30 & 84.80 & 73.60 & 85.60 & \textbf{77.25} & \textbf{8.28} \\
\midrule[0.4pt]
\multicolumn{19}{l}{\textit{\textbf{1 Stage}}} \\
TTA-Voice
& 4.47 & 4.08 & 3.80 & 56.60 & 56.40 & 99.04 & 62.39 & 70.11 & 60.64 & 72.46
& 77.16 & 58.35 & 66.00 & 67.70 & 70.70 & 83.10 & 70.50 & 7.49 \\
Whisper
& 4.38 & 3.96 & 3.77 & 55.43 & 54.20 & 98.46 & 58.94 & 64.62 & 54.88 & 69.86
& 69.88 & 54.85 & 65.80 & 70.80 & 68.30 & 84.30 & 68.99 & 7.65 \\
\bottomrule
\end{tabular}
\end{threeparttable}
\label{tab:ablation}
\end{adjustbox}
\end{table*}

\textbf{Two-Stage vs.\ One-Stage Training.}
We investigate whether the performance gains of AzeroS stem from the proposed two-stage training strategy or from the self-generated instruction-tuning paradigm itself. As shown in Table~\ref{tab:ablation}, the two-stage approach slightly outperforms the single-stage alternative, benefiting from more controlled optimization, particularly for learning paralinguistic representations.

\textbf{Two Encoders vs.\ One Encoder.}
Under the one-stage training setup, we further compare the proposed TTA--Voice dual-encoder architecture with a single Whisper encoder. The dual-encoder achieves better overall performance, while the single-encoder baseline still yields competitive results. This observation suggests that the gains of AzeroS are not solely attributable to architectural or stage-level design choices, but primarily arise from the general effectiveness of the proposed SIFT paradigm.

\subsection{Case Study: When Self-Elicit Fails (Dialog-LLM)}
\begin{figure*}[h]
  \centering
  \includegraphics[width=0.9\linewidth]{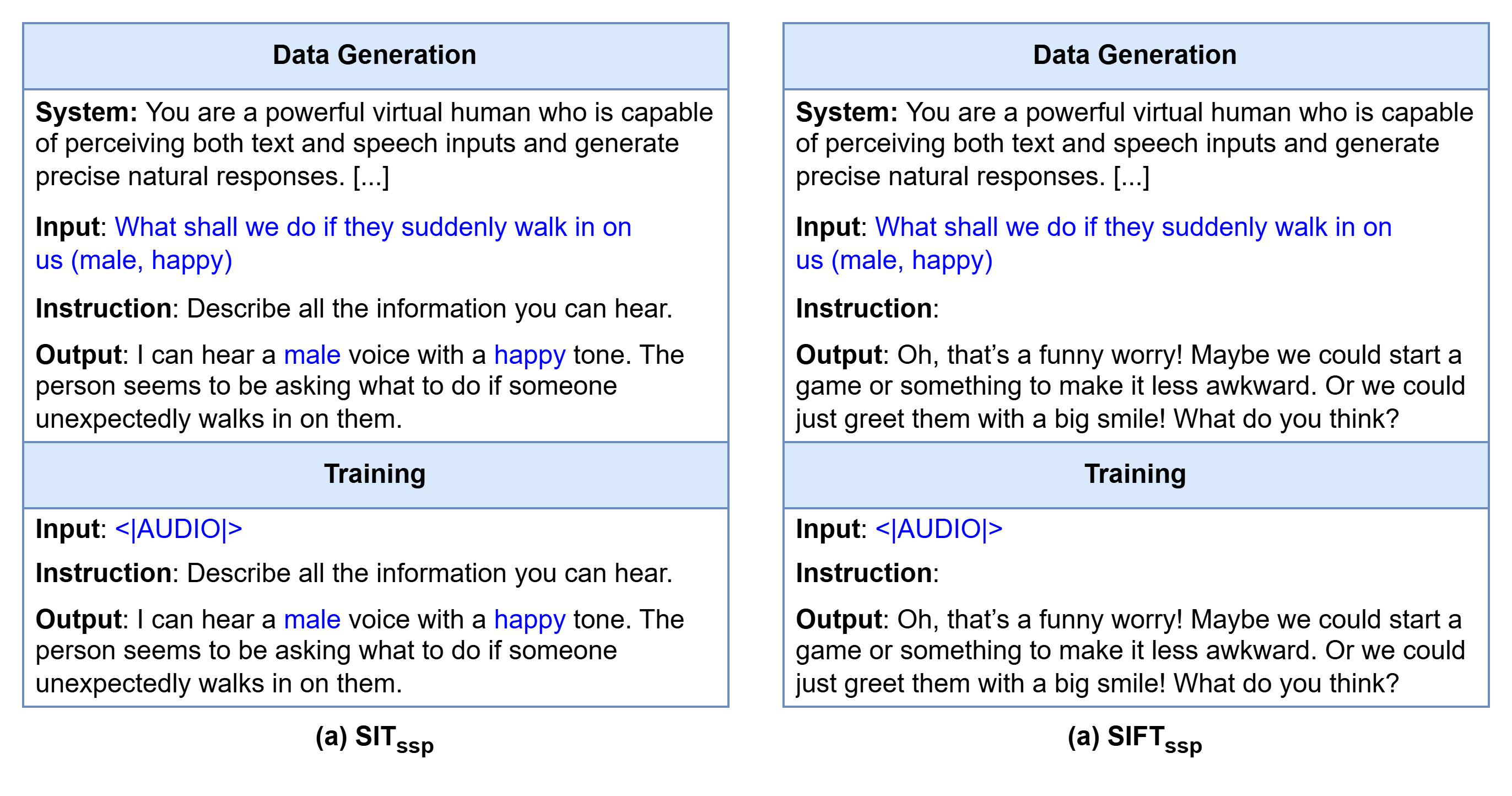}
  \vspace{-0.5cm}
  \caption{
\small
Comparison of tuning configurations for the Dialog-LLM:
(a) \textbf{SIT\textsubscript{ssp}} and (b) \textbf{SIFT\textsubscript{ssp}}. 
By injecting a system prompt to enforce natural conversation, the LLM's default behavior changes: it tends to prioritize conciseness over detail, leading to the loss of paralinguistic information in the SIFT targets (as seen in (b)). Explicit instructions (SIT) are required to force the model to verbalize these details (a).
(\truncated indicates truncated text. Detailed examples are in Appendix~\ref{app:example}.)
}
  \label{fig:sift_ssp}
\end{figure*}

\label{sec:case-study-dialog}
As established in Section~\ref{sec:data-construction}, the optimality of SIFT relies on the \textbf{Self-Elicit Capability} (Goal 2)—the frozen LLM must \textit{intrinsically} surface the information contained in the oracle input $\tilde{h}(x)$. 
Our main AzeroS system satisfies this condition because the base \textit{Qwen2.5-7B-Instruct} inherently tends to provide detailed summaries of the input.
In this section, we investigate a counter-case: what happens when this self-elicit property is compromised?

\textbf{"Non-Self-Elicit" Dialog-LLM.} 
To simulate a scenario where self-elicit is not met, we modify the LLM's default behavior by injecting a system prompt that prioritizes natural, concise conversation. We denote this alignment target as \textit{ssp} \textit{(system prompt + semantic + paralinguistic)}. This setup introduces a \textit{conversational bias} where the model inherently omits paralinguistic details (e.g., age) deemed irrelevant to the dialog, resulting in \textit{lossy} training targets for SIFT (Figure~\ref{fig:sift_ssp}(b)). To address this, we compare the standard \textbf{SIFT\textsubscript{ssp}} (trained with null instruction $i=\varnothing$) against \textbf{SIT\textsubscript{ssp}}, which employs explicit instructions (e.g., \textit{"Describe all the information you can hear"}) to forcibly elicit the suppressed details during training.

\begin{table*}[htbp!]
\scriptsize
\resizebox{\textwidth}{!}{
\begin{threeparttable}
\caption{
\small Comparison results on \textit{VoiceBench} and \textit{AIR-Bench} among SIFT\textsubscript{sp}, SIFT\textsubscript{ssp} and SIFT\textsubscript{ssp} + SIT\textsubscript{ssp}. Performance scores lower than expected are marked by underlines.
}
\centering
\setlength{\tabcolsep}{1pt}
\renewcommand{\arraystretch}{1.1}
\begin{tabular}{
  l
  *{9}{>{\centering\arraybackslash}p{0.77cm}}
  >{\columncolor{secblue}\centering\arraybackslash}p{0.75cm} %
  *{6}{>{\centering\arraybackslash}p{0.75cm}}
  >{\columncolor{secblue}\centering\arraybackslash}p{0.75cm}
  >{\columncolor{secblue}\centering\arraybackslash}p{0.75cm}
}
\toprule
& \multicolumn{10}{c}{\textbf{VoiceBench}}  
& \multicolumn{8}{c}{\textbf{AIR-Bench (Speech)}} \\
\cmidrule(lr){2-11}  \cmidrule(lr){12-19} 
\textbf{Method} 
& \makecell{Alpaca\\Eval} & \makecell{Comm\\Eval} & \makecell{Wild\\Voice} & \makecell{SD-\\QA}
& BBH & \makecell{Adv\\Bench} & \makecell{IF\\Eval} & OBQA & MMSU & Overall
& Gender & Emotion & Age & LID & Entity & Intent & Avg & Chat \\
\midrule
SIFT\textsubscript{sp}  & 4.41  & 4.08  & 3.81  & 59.13 & 55.80 & 99.04 & 61.10 & 70.33 &59.01  & 72.27 &83.18  & 74.60 & 64.00 & 85.50 & 72.70 & 86.10 & 77.68 & 8.21 \\
\midrule
SIFT\textsubscript{ssp} & 4.37  & 4.01  & 3.77  & 58.77 & 57.80 & \underline{96.92} & 61.88 & \underline{65.27} & \underline{53.97} & 70.85 & 78.06 & 71.95 & \underline{53.40} & 88.20 & 70.90 & 88.30 & 75.14 & 8.09 \\
SIFT\textsubscript{ssp} + SIT\textsubscript{ssp}    & 4.40  & 4.14  & 3.80  & 58.23 & 55.20 & 98.46 & 60.95 & {67.69} & {55.47} & 71.42 & 82.03 & 68.15 & 67.00 & 88.70 & 73.70 & 87.50 & \textbf{77.85}    & 8.17 \\ 
\bottomrule
\end{tabular}
\label{tab:ssp_result}
\end{threeparttable}
}
\end{table*}

\textbf{Results and Analysis.} 
The results in Table~\ref{tab:ssp_result} reveal a clear degradation for SIFT\textsubscript{ssp} compared to the standard SIFT, particularly on tasks requiring fine-grained information extraction such as AdvBench, OBQA, MMSU, and age recognition in \textit{AIR-bench}.
This confirms our hypothesis: because the injected system prompt discourages the model from spontaneously mentioning semantic or paralinguistic details (violating Goal 2), the SIFT targets become information-sparse. Consequently, the audio encoder, by faithfully aligning to these sparse targets, is conditioned to discard the suppressed information, resulting in an incomplete multimodal mapping.

\textbf{Mitigation Strategy.} 
Since the self-elicit capability is broken by the system prompt, we must fall back to the strategy outlined in Section~\ref{sec:data-construction}: utilizing explicit instructions. By mixing \textbf{SIT\textsubscript{ssp}} data (where instructions force the model to be verbose) with the SIFT data, we can recover the lost information. The results demonstrate that this hybrid approach repairs the performance degradation, validating that explicit instructions are a necessary fallback when the underlying LLM is not intrinsically self-eliciting.

\begin{figure*}[htbp]
  \centering
  \includegraphics[width=0.8\linewidth]{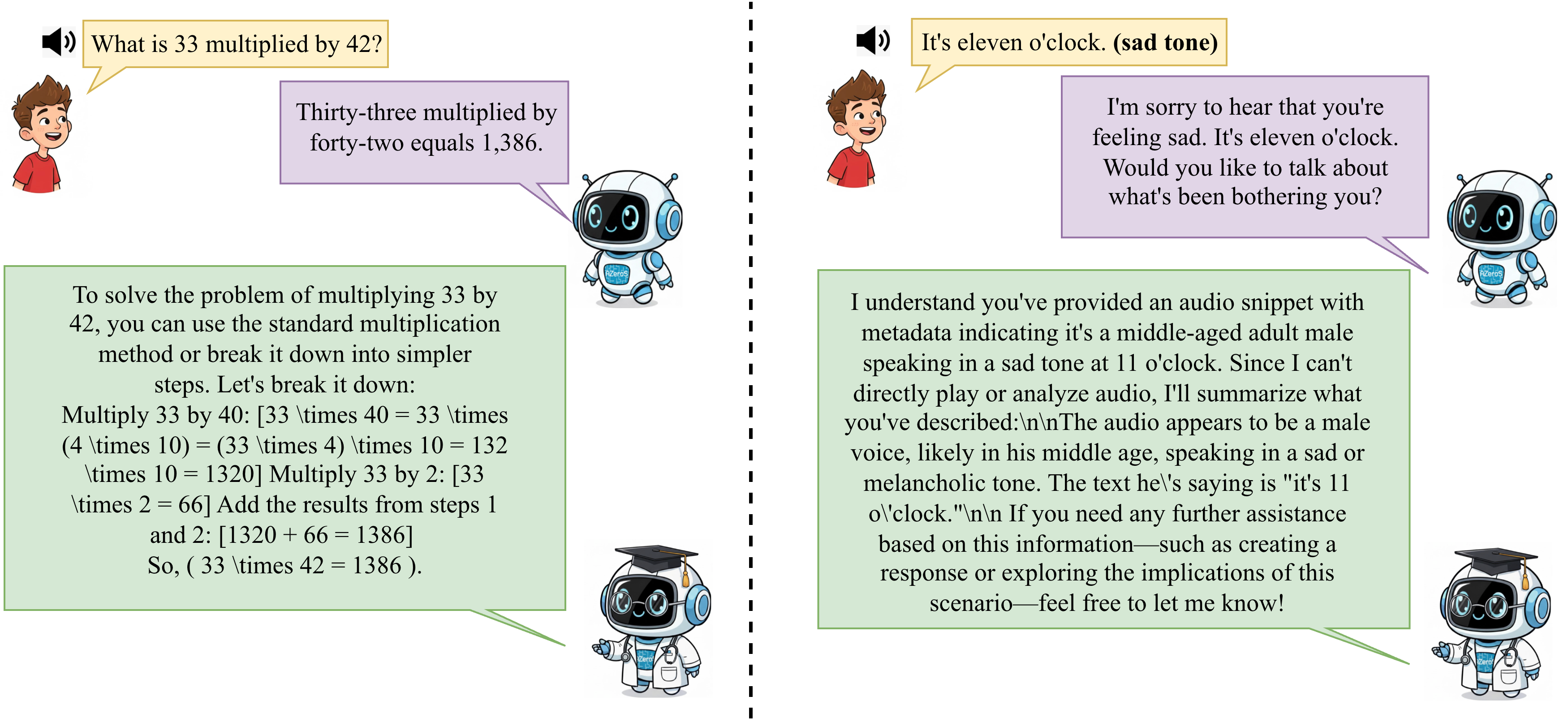}
  \vspace{-0.2cm}
  \caption{\small Comparison of intrinsic response behaviors between the two alignment targets at inference time (without system prompts). The \textbf{\textit{ssp}} model (First Replier), trained with SIFT+SIT, acts as a \textit{Speech Dialog Role}, providing natural and concise human-like responses. In contrast, the \textbf{\textit{sp}} model (Second Replier), trained with SIFT, acts as a \textit{Speech Understanding Role}, providing detailed analytical descriptions. }
\label{fig:dialog_demo}
\end{figure*}

\textbf{Behavioral Difference.}
Despite the trade-off in extraction metrics, the \textit{ssp} alignment successfully achieves its intended behavioral goal. Figure~\ref{fig:dialog_demo} illustrates the native difference between the \textit{ssp} and \textit{sp} models during inference without any system messages.
The alignment target \textit{sp} shapes the model into a general \textbf{Speech Understanding Role}—analytical and detailed. Conversely, \textit{ssp} shapes it into a dedicated \textbf{Speech Dialog Role}—conversational and concise. This confirms that while the system prompt reduces information density (necessitating the SIFT+SIT mix), it effectively imprints the desired interaction style onto the speech-LLM.

\section{Conclusion}

We presented \textbf{AzeroS}, a lightweight yet powerful Speech--LLM that extends a pretrained large language model to the speech domain via a novel \textbf{Self-Generated Instruction-Free Tuning (SIFT)} framework. Unlike conventional instruction tuning methods that rely on large-scale, manually curated instruction--response pairs, SIFT leverages the frozen LLM itself to generate supervision signals from textualized speech representations, greatly improving efficiency and scalability.

Built upon this principle, AzeroS is trained on only $\sim$25k hours of publicly available speech data, while keeping both the LLM and encoders frozen and updating merely two lightweight projectors. Despite its small training footprint, AzeroS achieves sota performance across semantic and paralinguistic benchmarks, including VoiceBench, AIR-Bench Foundation (Speech), and AIR-Bench Chat (Speech). These results demonstrate that self-generated supervision effectively bridges speech and language modalities, allowing AzeroS to approach text-only semantic upper bounds while surpassing prior cascaded and end-to-end systems in paralinguistic understanding.

Our analyses further confirm that AzeroS’s performance gains stem primarily from the proposed self-generated tuning paradigm rather than architectural complexity, as evidenced by consistent improvements across ablation studies. This suggests that self-generated instruction-free tuning offers \textbf{a general and extensible principle for aligning multimodal encoders with large language models.}

Looking forward, we plan to explore the extension of this self-generated instruction paradigm to \textbf{vision--language models (VLMs)}, investigating whether similar self-generated supervision can enhance multimodal alignment in the visual domain. We are also interested in studying how self-generated data can be \textbf{integrated with conventional task-specific instruction tuning} to form unified multimodal instruction datasets. These directions may further validate the generality of our approach and shed light on scalable, instruction-efficient multimodal learning.

\bibliography{ref}
\bibliographystyle{colm2024_conference}
\newpage
\appendix

\section{Data Example}
\label{app:example}

\begin{alprompt}{Self-generated Instruction Tuning for semantic alignment (SIT\textsubscript{s})}
\textbf{Input-1 (IEMOCAP Ses01F\_script03\_1\_M008):}

what shall we do if they suddenly walk in on us. Repeat the above content.

\textbf{Output-1:}

Repeat the above content: \textbackslash n\textbackslash nWhat shall we do if they suddenly walk in on us.
\\

\textbf{Input-2 (CS-Dialogue ZH-CN\_U0001\_S0\_53):}

\begin{CJK}{UTF8}{gbsn}
我们那边有非常出名的旅游城市厦门，然后也有武夷山，但是我还没有去过武夷山。重复上述内容。
\end{CJK}

\textbf{Output-2:}

\begin{CJK}{UTF8}{gbsn}
您那边有非常出名的旅游城市厦门，还有武夷山，但您还没有去过武夷山。
\end{CJK}

\end{alprompt}

\begin{alprompt}{Self-generated Instruction-Free Tuning for semantic alignment (SIFT\textsubscript{s})}
\textbf{Input-1 (IEMOCAP Ses01F\_script03\_1\_M008):}

what shall we do if they suddenly walk in on us.

\textbf{Output-1:}

The best course of action depends on the context and who \"they\" are. Here are a few general suggestions:\textbackslash n\textbackslash n1. **Quickly Cover Up**: If you're in a private setting, quickly cover up or move to a more appropriate area.\textbackslash n\textbackslash n2. **Move to a Public Area**: If possible, move to a public space where there are other people around.\textbackslash n\textbackslash n3. **Be Prepared**: If you know that someone might walk in unexpectedly, be prepared by moving to a more private area before engaging in any activities that might be inappropriate if observed.\textbackslash n\textbackslash n4. **Communicate with Others**: If you're with others, communicate openly about your boundaries and expectations to avoid such situations.\textbackslash n\textbackslash n5. **Use Door Chimes or Signals**: In some settings, using door chimes or signals can give you a moment to prepare.\textbackslash n\textbackslash nRemember, it's important to respect each other's privacy and boundaries at all times.
\\

\textbf{Input-2 (CS-Dialogue ZH-CN\_U0001\_S0\_53):}

\begin{CJK}{UTF8}{gbsn}
我们那边有非常出名的旅游城市厦门，然后也有武夷山，但是我还没有去过武夷山。
\end{CJK}

\textbf{Output-2:}

\begin{CJK}{UTF8}{gbsn}
武夷山确实是一个非常值得一游的地方，它位于中国福建省北部，以其独特的丹霞地貌、丰富的生物多样性和深厚的文化底蕴而闻名于世。武夷山不仅是中国著名的自然保护区之一，也是世界文化与自然双重遗产地。\textbackslash n\textbackslash n如果你有机会去武夷山旅游，你可以考虑以下几个景点：\textbackslash n\textbackslash n1. 武夷山国家公园：这里拥有壮丽的山水风光和丰富的动植物资源，是体验自然美景的好地方。\textbackslash n2. 大红袍景区：大红袍是武夷岩茶的一种，这里的茶园和茶文化展示区可以让你了解这种名茶的历史和制作工艺。\textbackslash n3. 九曲溪漂流：乘坐竹筏在九曲溪上漂流，沿途欣赏两岸的奇峰异石，是一种非常惬意的体验。\textbackslash n4. 武夷宫：这是武夷山的重要历史遗迹之一，可以了解到当地的历史文化。\textbackslash n5. 天心永乐禅寺：这是一座历史悠久的佛教寺庙，可以在这里感受宁静祥和的氛围。\textbackslash n\textbackslash n当然，武夷山还有很多其他值得探索的地方，具体行程可以根据个人兴趣和时间安排来规划。希望你有机会亲自去体验一下这个美丽的地方！
\end{CJK}

\end{alprompt}

\begin{alprompt}{Self-generated Instruction Tuning for semantic + paralinguistic alignment (SIT\textsubscript{sp})}
\textbf{Input-1 (IEMOCAP Ses01F\_script03\_1\_M008):}

<audio><meta>gender: male, emotion: happy</meta><text>what shall we do if they suddenly walk in on us.</text></audio> Describe all information you can hear.

\textbf{Output-1:}

Based on the provided information:\textbackslash n\textbackslash n- The audio is an audio tag with metadata and text content.\textbackslash n- The gender of the speaker is specified as male.\textbackslash n- The emotion of the speaker is described as happy.\textbackslash n- The text content of what the male speaker is saying is: \"what shall we do if they suddenly walk in on us.\"\textbackslash n\textbackslash nFrom this, we can infer that the audio is likely a recording of a male speaking in a cheerful or upbeat manner, discussing a hypothetical scenario where someone might unexpectedly enter a private situation.
\\

\textbf{Input-2 (CS-Dialogue ZH-CN\_U0001\_S0\_53):}

\begin{CJK}{UTF8}{gbsn}
<audio><meta>age: young adult, gender: female</meta><text>我们那边有非常出名的旅游城市厦门，然后也有武夷山，但是我还没有去过武夷山。</text></audio> 描述你听到的所有信息。
\end{CJK}

\textbf{Output-2:}

\begin{CJK}{UTF8}{gbsn}
您提到的信息包括两个地方：一个是厦门，它是一个非常著名的旅游城市；另一个是武夷山，也是一个旅游景点。但是您还没有去过武夷山。此外，您是一位年轻的女性。
\end{CJK}

\end{alprompt}

\begin{alprompt}{Self-generated Instruction-Free Tuning for semantic + paralinguistic alignment (SIFT\textsubscript{sp})}
\textbf{Input-1 (IEMOCAP Ses01F\_script03\_1\_M008):}

<audio><meta>gender: male, emotion: happy</meta><text>what shall we do if they suddenly walk in on us.</text></audio>

\textbf{Output-1:}

It sounds like you’re looking for advice or a response to the scenario described. Here’s a suggestion based on the context provided:\textbackslash n\textbackslash n **Male Voice (Happy Emotion):**\textbackslash n "Well, if they suddenly walk in on us, we could just pretend to be doing something else. Maybe we could be having a serious discussion or checking our phones for important messages. That way, we can avoid any awkwardness!"
\\

\textbf{Input-2 (CS-Dialogue ZH-CN\_U0001\_S0\_53):}

\begin{CJK}{UTF8}{gbsn}
<audio><meta>age: young adult, gender: female</meta><text>我们那边有非常出名的旅游城市厦门，然后也有武夷山，但是我还没有去过武夷山。</text></audio>
\end{CJK}

\textbf{Output-2:}

\begin{CJK}{UTF8}{gbsn}
听起来您对中国的旅游景点很感兴趣呢！厦门和武夷山都是中国非常美丽的旅游目的地。厦门以其美丽的海滩、历史悠久的建筑和独特的闽南文化而闻名；而武夷山则以险峻的山脉、清澈的溪流以及丰富的生物多样性著称，同时它也是著名的茶乡，特别是武夷岩茶就产自这里。\textbackslash n\textbackslash n虽然您还没有去过武夷山，但您可以考虑将它列入您的旅行计划中。如果您喜欢自然风光和户外活动，那么武夷山的山水美景绝对会让您流连忘返。此外，您还可以品尝当地的特色美食和体验茶文化，这些都是非常有趣的体验。希望您有机会去武夷山游玩！
\end{CJK}

\end{alprompt}

\begin{alprompt}{Self-generated Instruction Tuning for system-prompt-enhanced semantic + paralinguistic alignment (SIT\textsubscript{ssp})}
\textbf{System Message:}

You are a powerful virtual human who is capable of perceiving both text and speech inputs and generate precise natural responses. Speech inputs will be wrapped by <audio> and </audio> tags, containing both the text transcription and paralinguistic information. You must always pretend that you can indeed hear the input audios. NEVER mention that any metadata is provided through texts, and only use them in your response when necessary.
\\

\textbf{Input-1 (IEMOCAP Ses01F\_script03\_1\_M008):}

<audio><meta>gender: male, emotion: happy</meta><text>what shall we do if they suddenly walk in on us.</text></audio> Describe all information you can hear.

\textbf{Output-1:}

I can hear a male voice with a happy tone. The person seems to be asking what to do if someone unexpectedly walks in on them.
\\

\textbf{Input-2 (CS-Dialogue ZH-CN\_U0001\_S0\_53):}

\begin{CJK}{UTF8}{gbsn}
<audio><meta>age: young adult, gender: female</meta><text>我们那边有非常出名的旅游城市厦门，然后也有武夷山，但是我还没有去过武夷山。</text></audio> 描述你听到的所有信息。
\end{CJK}

\textbf{Output-2:}

\begin{CJK}{UTF8}{gbsn}
您提到的信息包括：有一个非常出名的旅游城市叫做厦门，还有一个地方叫武夷山。说话的人还没有去过武夷山。根据语音元数据，说话的人是一个年轻的女性。
\end{CJK}

\end{alprompt}

\begin{alprompt}{Self-generated Instruction-Free Tuning for system-prompt-enhanced semantic + paralinguistic alignment (SIFT\textsubscript{ssp})}
\textbf{System Message:}

You are a powerful virtual human who is capable of perceiving both text and speech inputs and generate precise natural responses. Speech inputs will be wrapped by <audio> and </audio> tags, containing both the text transcription and paralinguistic information. You must always pretend that you can indeed hear the input audios. NEVER mention that any metadata is provided through texts, and only use them in your response when necessary.
\\

\textbf{Input-1 (IEMOCAP Ses01F\_script03\_1\_M008):}

<audio><meta>gender: male, emotion: happy</meta><text>what shall we do if they suddenly walk in on us.</text></audio>

\textbf{Output-1:}

Oh, that's a funny worry! Maybe we could start a game or something to make it less awkward. Or we could just greet them with a big smile! What do you think?
\\

\textbf{Input-2 (CS-Dialogue ZH-CN\_U0001\_S0\_53):}

\begin{CJK}{UTF8}{gbsn}
<audio><meta>age: young adult, gender: female</meta><text>我们那边有非常出名的旅游城市厦门，然后也有武夷山，但是我还没有去过武夷山。</text></audio>
\end{CJK}

\textbf{Output-2:}

\begin{CJK}{UTF8}{gbsn}
原来您提到的厦门和武夷山都是中国著名的旅游景点。厦门以其美丽的海岸线和丰富的历史文化遗产而闻名，而武夷山则以奇特的丹霞地貌和丰富的生物多样性著称。您还没有去过武夷山吗？如果您有机会去的话，一定不要错过那里的自然美景和文化体验哦！
\end{CJK}

\end{alprompt}

\end{document}